# Comment on "Fastest learning in small-world neural networks"

Z.X. Guo

*Institute of Textiles and Clothing, The Hong Kong Polytechnic University, Hong Kong, China*

## Abstract

This comment reexamines Simard et al.'s work in [D. Simard, L. Nadeau, H. Kröger, Phys. Lett. A 336 (2005) 8-15]. We found that Simard et al. calculated mistakenly the local connectivity lengths $D_{local}$ of networks. The right results of $D_{local}$ are presented and the supervised learning performance of feedforward neural networks (FNNs) with different rewirings are re-investigated in this comment. This comment discredits Simard et al's work by two conclusions: 1) Rewiring connections of FNNs cannot generate networks with small-world connectivity; 2) For different training sets, there do not exist networks with a certain number of rewirings generating reduced learning errors than networks with other numbers of rewiring.

* Corresponding author. Tel.: +855 27664024, Fax: +855 27664024
E-mail address: tcguozx@inet.polyu.edu.hk.

# 1. Introduction

Simard et al. [1] claimed that networks with small-world connectivity can be constructed by rewiring some connections of feed-forward neural networks (FNNs), which give less learning errors than the networks of regular or random connectivity. In [1], the small-world network architecture is measured by small global and local connectivity lengths $D_{global}$ and $D_{local}$, which are defined via the concept of global and local efficiency $E_{global}$ and $E_{local}$ [2,3]. $E_{local}$ is defined as the average efficiency of subgraphs. Subgraph $G_i$ of neighbors of neuron $i$ is formed by the neurons directly connected to neuron $i$ according to the definition in [3,4]. However, in [1], all neurons occurring in the same layer as neuron $i$ are also included in $G_i$. That is, $G_i$ is mistakenly defined in [1]. The conclusion in [1], that the small-world network can be constructed by randomly rewiring the connections of FNNs, is thus questionable.

In [1], the learning performance of the network with a certain number of rewired connections is observed based on one training set and one random network connectivity. However, different training sets and different network connectivities can generate different learning performances, by which different conclusions can be drawn.

In this comment, we reinvestigate the values of $D_{global}$ and $D_{local}$ of FNNs with different numbers of rewired connections, and the supervised learning performance of these networks.

## 2. Network connectivity lengths

In this comment, the $G_i$ is formed by the neurons directly connected to neuron $i$ according to the definition in [3,4]. We investigate the relations between $D_{local}$ and the number of rewirings in terms of the following 4 FNNs:

Network A: a network of 5 neurons per layer and 5 layers

Network B: a network of 5 neurons per layer and 8 layers

Network C: a network of 15 neurons per layer and 8 layers

Network D: a network of 10 neurons per layer and 10 layers

Given a specified number of rewired connections, different network connectivities can be obtained by randomly cutting and rewiring connections. We use 100 different connectivities generated randomly to compute $D_{local}$ and $D_{global}$ of a specified number of rewirings. Figure 1 shows $D_{local}$ and $D_{global}$ as a function of the number of rewired connections for the 4 networks, in which figures 1.(a)-(d) represents the results of networks A - D respectively. It is evident that the regime of small-world architecture does not exist for any of the 4 networks. That is, rewiring some connections of FNNs cannot (at least for networks A-D) form networks with small-world connectivity although previous studies revealed that the small-world network can be obtained by randomly rewiring some connections of regular lattice [3,4]. Simard et al. make a wrong conclusion in [1] because they mistakenly compute $D_{local}$ and $D_{global}$.

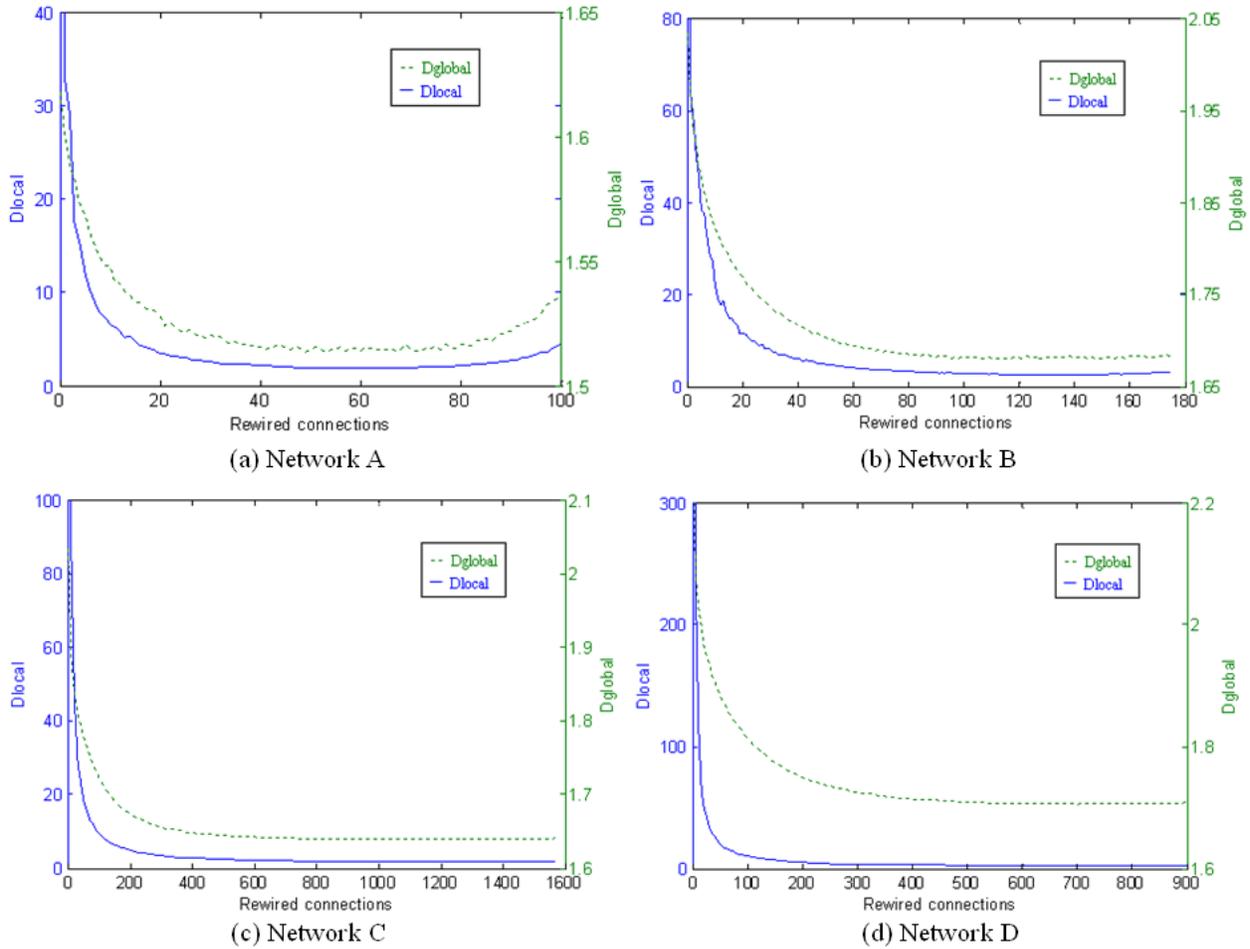

Fig. 1. $D_{local}$ and $D_{global}$ versus number of rewired connections.

## 2. Supervised learning

In this section, we reexamine whether the FNN with some random rewirings gives reduced learning errors. The supervised learning performances of networks B, C and D are investigated by training them with random binary input and output patterns (training set):

Except for different training sets and network connectivities, the networks are trained based on the same network setting presented in [1]. The learning algorithm is back-propagation. For the 3 networks, the relationships between learning error and the number of rewirings are shown in Fig.s 2-4 respectively. In each figure, (a) and (b) show the results of 150000 and 300000 iterations respectively. '□' and '■' represent the minimum and the mean of mean absolute errors (MAEs) of multiple tests respectively. For networks B-D, their smallest means of MAEs are generated, respectively, by the networks at $N_{rewire}$ =90, 1100, and 750. These results are quite inconsistent with the results in [1], in which $N_{rewire}$ equals 28, 830, 400 respectively.

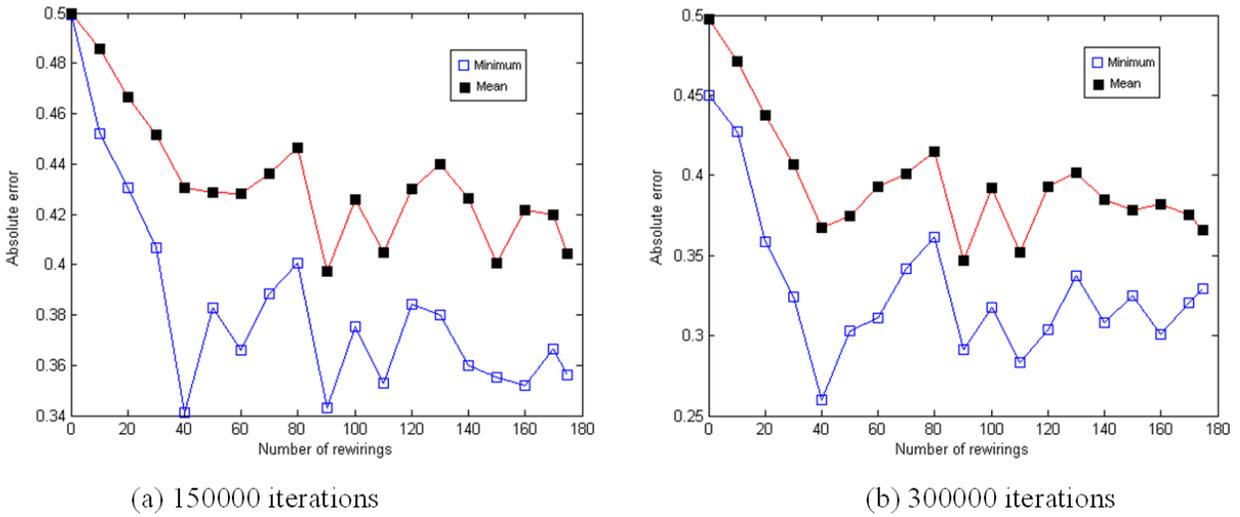

Fig. 2.   Learning results of network B. Learning of 40 patterns, learning rate 0.01, 20 statistical tests.

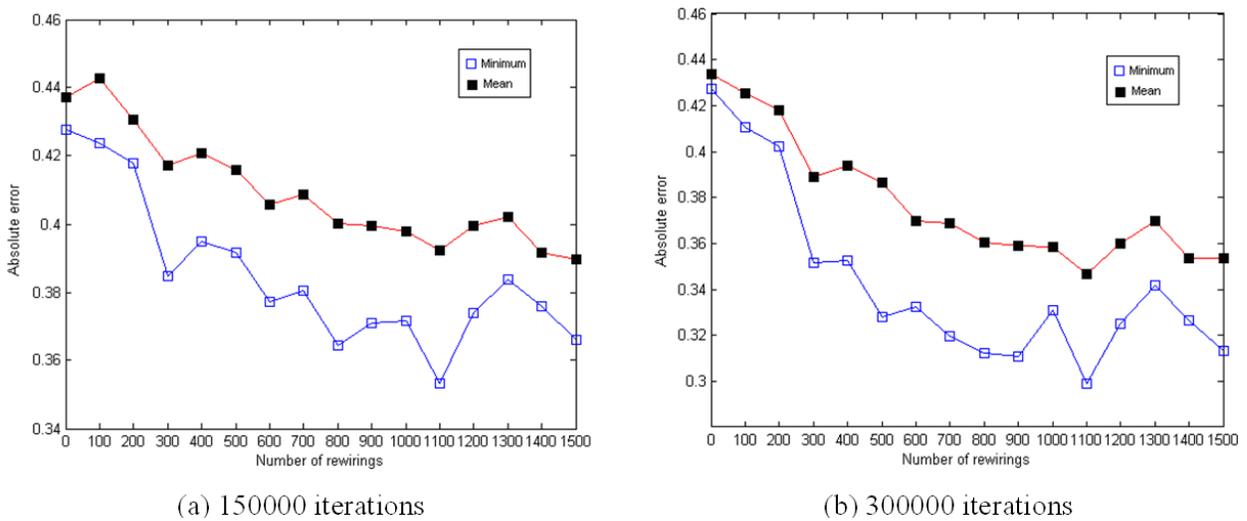

Fig. 3.   Learning results of network C. Learning of 40 patterns, learning rate 0.01, 17 statistical tests.

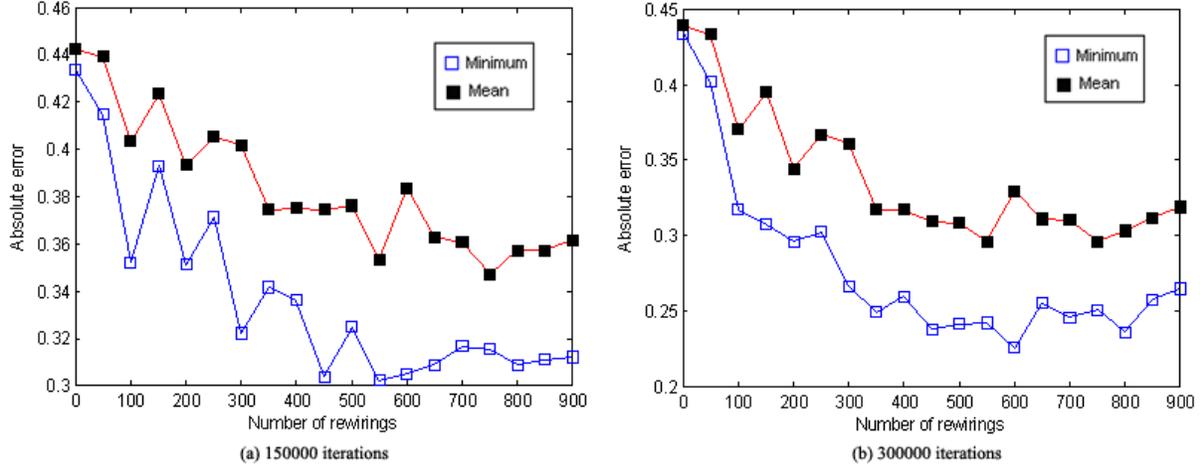

Fig. 4.  Learning results of network D. Learning of 30 patterns, learning rate 0.02, 20 statistical tests.

Next, we examine further whether the network with a certain $N_{rewire}$ (like in the regime of small-world architecture) can generate reduced learning errors than networks with other number of rewiring. For a specified $N_{rewire}$, different network connectivities can be generated. Taking network D as an example, its learning performance is investigated further based on 4 cases in terms of different connectivities and training sets. The learning results of 300000 iterations are shown in Fig. 5. It is clear that the best learning performances are obtained, for different cases, at quite different number of rewirings. That is, for different training sets, there do not exist networks with a certain $N_{rewire}$ capable of generating less learning errors than networks with other $N_{rewire}$, which is also inconsistent with the conclusion in [1].

In [1], Simard et al. claimed that the networks at a certain $N_{rewire}$ give reduced learning errors based on only one training set and one network connectivity. Our experimental results discredited this conclusion because different conclusions were drawn in terms of more training sets and network connectivities.

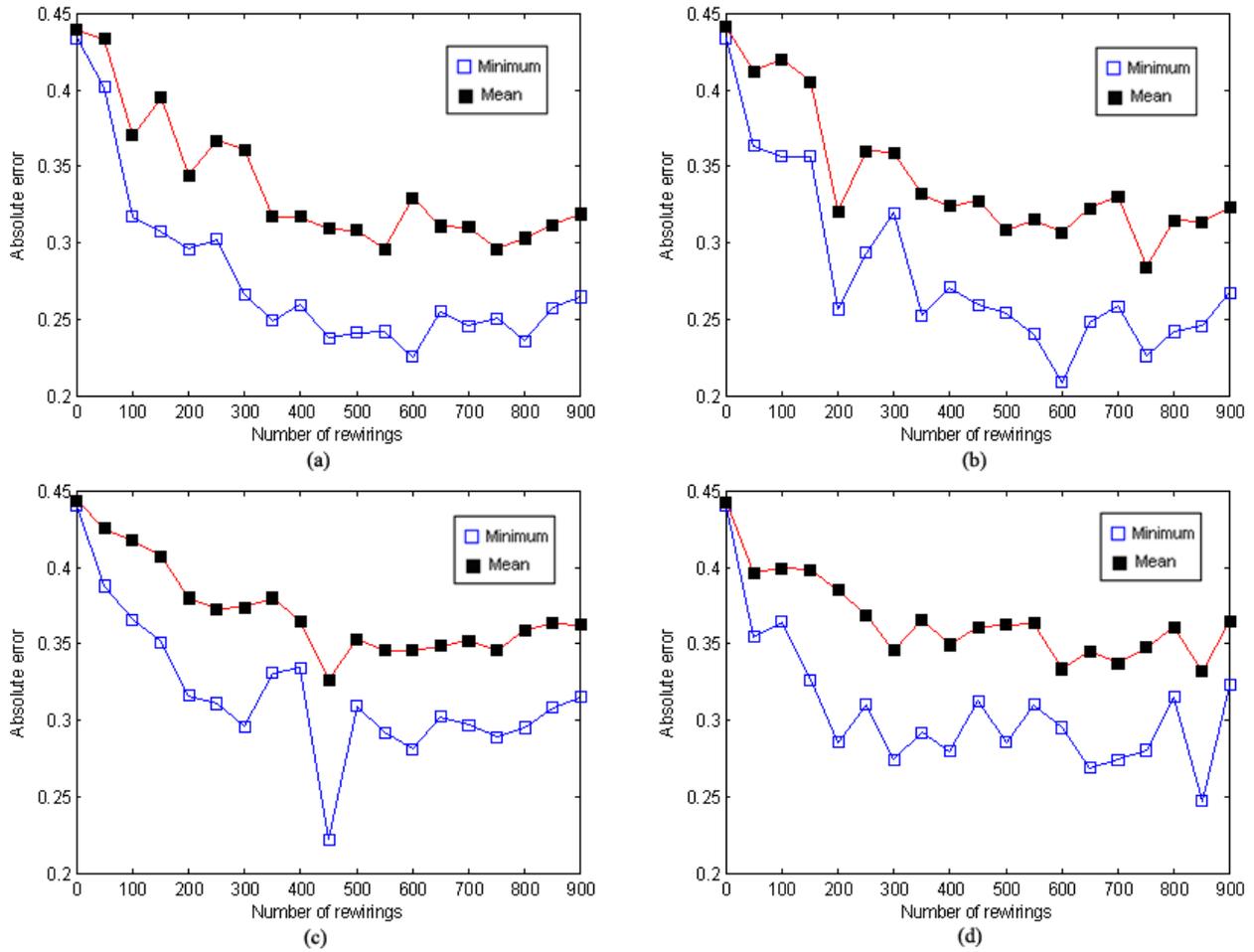

Fig. 5. Learning results of network D based on different training sets and network connectivities. (a) results based on training set 1 and connectivity 1; (b) results based on training set 1 and connectivity 2; (c) results based on training set 2 and connectivity 2; (d) results based on training set 2 and connectivity 3.

## 3. Conclusions

This comment reexamines Simard et al.'s work [1] by re-calculating $D_{local}$ of networks and re-investigating their supervised learning performance. Two important conclusions can be drawn from experimental results that: 1) Rewiring randomly some connections of FNNs cannot construct small-world networks; 2) For different training sets, results generated by networks with some random rewirings are superior to those generated by regular FNNs. However, there do not exist networks with a certain $N_{rewire}$ generating less learning errors than networks with other $N_{rewire}$. These conclusions discredit Simard et al's conclusions in [1].